\documentclass[11pt,a4paper]{article}
\usepackage[T1]{fontenc}
\usepackage[margin=1in]{geometry}
\makeatletter
\newcommand{\@articletype}{}
\newcommand{\articletype}[1]{\gdef\@articletype{#1}}
\newcommand{\name}[1]{{\large #1\par}\vspace{0.6em}}
\newcommand{\affil}[1]{{\small #1\par}\vspace{0.3em}}
\renewcommand{\@maketitle}{%
  \newpage
  \null
  \vskip 1em
  \begin{center}%
    {\small \@articletype\par}%
    \vskip 0.8em
    {\LARGE\bfseries \@title\par}%
    \vskip 1.5em
    {\@author\par}%
  \end{center}%
  \par
  \vskip 1.5em
}
\makeatother
\date{}
\newenvironment{keywords}
  {\par\medskip\noindent\textbf{Keywords: }\ignorespaces}
  {\par\medskip}
\usepackage{lmodern}
\usepackage{amsmath,amssymb,amsthm}
\usepackage{graphicx}
\usepackage{xcolor}
\usepackage{array}
\definecolor{revisionblue}{RGB}{0,0,255}

\usepackage{booktabs}
\usepackage{algorithm}
\usepackage{algpseudocode}
\usepackage[numbers,sort&compress]{natbib}
\bibpunct[, ]{[}{]}{,}{n}{,}{,}
\renewcommand\bibfont{\fontsize{10}{12}\selectfont}
\usepackage{xurl}
\usepackage[hidelinks]{hyperref}
\usepackage{orcidlink}
\theoremstyle{plain}
\newtheorem{theorem}{Theorem}[section]
\newtheorem{proposition}[theorem]{Proposition}

\begin{document}

\articletype{Research Article}
\title{Hybrid Neural Network Training with Two-Dimensional Model-Based Subspace Search}
\author{%
\name{Churui Wang\orcidlink{0009-0003-0461-5969} \textsuperscript{a} and Pengcheng Xie\,\orcidlink{0000-0001-5973-1535}\textsuperscript{b,*}%
\thanks{*Corresponding author. Email: \protect\href{mailto:pxie@lbl.gov}{pxie@lbl.gov}}}
\affil{\textsuperscript{a}University of California, Berkeley, Berkeley, CA, USA}
\affil{\textsuperscript{b}Applied Mathematics and Computational Research Division, Lawrence Berkeley National Laboratory, Berkeley, CA, USA}
}
\maketitle

\begin{abstract}
Training neural networks without derivatives is challenging because full-dimensional quadratic models require a number of coefficients that grows quadratically with the parameter dimension. We investigate \texttt{2D-MoSub}, which constructs quadratic models in two-dimensional subspaces, as both a standalone optimizer and a complement to Adam. Two hybrid strategies are considered: periodic subspace interventions and an Adam phase followed by derivative-free refinement. In the reported noisy Ackley fitting experiment with 1,341 parameters, standalone \texttt{2D-MoSub} reaches a lower mean squared error than finite-difference Adam. In a flow-matching experiment with 12,866 parameters, standalone search stagnates, whereas Adam and the hybrids reduce the loss; the available curves do not establish a consistent advantage of the hybrids over Adam. We analyze conditional unbiasedness and noise amplification for fixed-geometry stochastic quadratic models, characterize the limited guarantee provided by exact-loss acceptance, and derive stationarity bounds for randomized and deterministic hybrids with stochastic gradient steps. These bounds do not apply directly to Adam. The analysis also identifies the endpoint regularization needed for the reported flow-matching path and includes the cost of model construction and acceptance checks. Together, the results define conditions under which subspace training can be assessed without equating a small interpolation problem with dimension-independent training cost.
\end{abstract}


\section{Introduction}
\label{sec:introduction}

\subsection{Background and motivation}
Optimization is central to scientific computing, machine learning, and engineering. In applications involving simulations, legacy software, or restricted interfaces, an objective may be evaluable even when its derivatives are unavailable or unreliable. Derivative-free optimization (DFO) addresses this setting by using function values to guide the search; standard references include \cite{conn2009introduction,audet2017derivative,larson_menickelly_wild_2019,zhang2021}.

Model-based trust-region methods construct local approximations of the objective and minimize them within a region in which the approximation is expected to be useful. Quadratic models can capture local curvature without explicit derivatives \cite{Winfield1969,Powell2003,Conn2009a,Gratton-2017,cartis2019improving}. Developments in model construction include least-norm quadratic models \cite{xieyuannew}, least $H^2$-norm updates \cite{10.1093/imanum/drae106}, and regional minimal updating \cite{xie2025remuregionalminimalupdating}. Related work studies convergence and distance reduction \cite{XIE2025116146}, objective transformations \cite{xie2023dfoto}, and privacy-preserving black-box optimization \cite{xie2025privacypreservingblackboxoptimizationpbbo}.

These methods face a substantial dimensionality barrier. A general quadratic in $n$ variables has $(n+1)(n+2)/2$ coefficients, so determining it by interpolation requires a number of suitably positioned points that grows quadratically with $n$. Underdetermined models can reduce this requirement, but their construction and interpolation geometry remain important considerations. Subspace methods instead restrict individual searches to a lower-dimensional space \cite{yuan2007subspace,yuan2009subspace,yuan2014review,cartis2023scalable}.

\subsection{Subspace methods and neural network training}
Existing subspace approaches differ in how they select directions and maintain their models. The method described in \cite[Chapter 5.3]{zhang012} and \cite{zhang2025scalablederivativefreeoptimizationalgorithms} combines two-dimensional searches with full-dimensional finite differences. Other approaches update a full-dimensional model using random subspaces \cite{Menickelly2023}, or construct fully quadratic models in randomized subspaces \cite{cartis2024randomizedsubspacederivativefreeoptimization}. Subspace techniques also arise in gradient-based methods \cite{zhang2023drsomdimensionreducedsecondorder}.

This paper builds on \texttt{2D-MoSub} \cite{xie2023twodimensional}. The method combines a direction informed by previous iterations with an orthogonal exploration direction and constructs a quadratic model in the resulting two-dimensional subspace. A general quadratic in this subspace has only six coefficients, irrespective of the ambient dimension. This reduces the size of the model-building problem, although each objective evaluation and each parameter update may still be expensive for a large network.

For neural network training, low-dimensional searches offer both opportunities and limitations. They can exploit local curvature using only loss evaluations, but a selected subspace may capture little useful descent information, particularly at initialization. When gradients are available, an initial gradient-based phase or periodic gradient updates may help identify regions in which subsequent subspace searches are more productive. This motivates studying \texttt{2D-MoSub} both as a standalone derivative-free optimizer and as a component of hybrid training. The hybrid methods require gradient access during their Adam updates.

\subsection{Scope and contributions}
We investigate these strategies on two tasks with different parameter dimensions. The first fits noisy samples of the Ackley function with a network containing 1,341 trainable parameters. The second learns a vector field for a flow-matching model with 12,866 trainable parameters. These examples allow us to examine function-evaluation efficiency, the limitations of standalone subspace search, and the effect of combining subspace steps with Adam.

The contributions are threefold. First, we formulate interleaved and sequential strategies for incorporating \texttt{2D-MoSub} into neural network training, with an explicit distinction between model-building losses and the objective used for acceptance. Second, we assess these strategies on noisy regression and flow matching, visualizing their performance and ensuring that these strategies do improve the final quality. Third, we establish fixed-geometry model unbiasedness and a covariance formula, and derive stationarity bounds for safeguarded hybrids with stochastic gradient steps under randomized and deterministic schedules. These results clarify the roles of sampling, interpolation geometry, and acceptance cost.

Section~\ref{sec:hybrid} describes the training strategies. Section~\ref{section:Numerical results} presents the experiments, and Section~\ref{sec:theory} provides the analysis. Section~\ref{sec:conclusion} discusses the conclusions and remaining questions. Appendix~\ref{section:2D-MoSub} summarizes the underlying subspace method; implementation details are available in \cite{xie2023twodimensional}.

\section{Hybrid training with \texttt{2D-MoSub}}
\label{sec:hybrid}
\subsection{Subspace models and update schedules}
Let $\boldsymbol{\theta}\in\mathbb{R}^n$ denote the network parameters and let $\mathcal{L}(\boldsymbol{\theta})$ be the training objective. At a subspace iteration, an orthonormal matrix $\boldsymbol{P}\in\mathbb{R}^{n\times 2}$ defines the affine search space $\boldsymbol{\theta}+\operatorname{range}(\boldsymbol{P})$. The method approximates $\boldsymbol{y}\mapsto\mathcal{L}(\boldsymbol{\theta}+\boldsymbol{P}\boldsymbol{y})$ by a quadratic model in $\boldsymbol{y}\in\mathbb{R}^2$ and computes a trial step within a trust region. Retaining a direction from earlier iterations allows information to be reused, while the second direction provides exploration outside that direction.

We consider two ways to combine these searches with Adam. In \emph{interleaved training}, one cycle contains $q-1$ Adam updates followed by one subspace intervention, with integer $q\ge2$. In \emph{sequential training}, Adam is used for $K_{\mathrm{A}}$ updates and subspace search for the remaining $K-K_{\mathrm{A}}$ interventions. Writing a subspace model as
\begin{equation}
\label{eq:subspace-model}
 m_k(\boldsymbol{y})=c_k+\boldsymbol{g}_k^T\boldsymbol{y}+\tfrac12 \boldsymbol{y}^T\boldsymbol{H}_k\boldsymbol{y},
 \qquad \boldsymbol{y}\in\mathbb{R}^2,\quad \boldsymbol{H}_k=\boldsymbol{H}_k^T,
\end{equation}
gives six scalar coefficients. A trial coordinate approximately minimizes $m_k$ over $\|\boldsymbol{y}\|_2\le\Delta_k$, and produces $\boldsymbol{\theta}_{\mathrm{trial}}=\boldsymbol{\theta}_k+\boldsymbol{P}_k\boldsymbol{y}$. Six suitably positioned values determine a general quadratic from scratch. Forming the directions and evaluating a network mainly depend on the ambient dimension $n$.

\subsection{Acceptance objective and optimizer state}
Algorithm~\ref{alg:interleaved} specifies a safeguarded version of interleaved training. Model construction uses a small batch, whereas acceptance compares the candidate and current parameters on a fixed empirical objective $\mathcal{L}$.

\begin{algorithm}[htbp]
\caption{Interleaved Adam and \texttt{2D-MoSub} training with exact-loss acceptance}
\label{alg:interleaved}
\begin{algorithmic}[1]
\Require Initial parameters $\boldsymbol{\theta}_0$, updates $K$, integer period $q\ge2$, batch sizes $B,b$, inner evaluation budget $r$, and fixed empirical objective $\mathcal{L}$.
\State Initialize Adam's moments.
\For{$k=0,\ldots,K-1$}
    \If{$k+1\not\equiv0\pmod q$}
        \State Sample a batch of size $B$; advance Adam's state and compute $\boldsymbol{\theta}_{k+1}$.
    \Else
        \State Sample a batch of size $b$ and retain it during model construction.
    \State Preserve $\boldsymbol{\theta}_k$; construct $\boldsymbol{\theta}_{\mathrm{trial}}$ using at most $r$ batch-loss evaluations.
        \State Evaluate $\mathcal{L}(\boldsymbol{\theta}_{\mathrm{trial}})$ and $\mathcal{L}(\boldsymbol{\theta}_k)$ on the fixed acceptance data.
        \If{$\mathcal{L}(\boldsymbol{\theta}_{\mathrm{trial}})<\mathcal{L}(\boldsymbol{\theta}_k)$}
            \State $\boldsymbol{\theta}_{k+1}\gets\boldsymbol{\theta}_{\mathrm{trial}}$.
        \Else
            \State $\boldsymbol{\theta}_{k+1}\gets\boldsymbol{\theta}_k$.
        \EndIf
    \EndIf
\EndFor
\State \Return $\boldsymbol{\theta}_K$.
\end{algorithmic}
\end{algorithm}

In simple terms, the safeguarded scheme retains Adam's moment estimates during a subspace intervention and updates with \texttt{2D-MoSub} intervention at the end of a circle. After an intervention, the next gradient is evaluated at the accepted parameters. This is a convention for Algorithm~\ref{alg:interleaved}; Rejected proposals must leave the network parameters at $\boldsymbol{\theta}_k$.

This acceptance rule prevents a subspace proposal from increasing the fixed empirical loss. It does not enforce monotonicity of the intervening Adam updates or of losses evaluated on fresh random batches. Its computational cost includes the full-objective comparisons as well as all evaluations used to construct and minimize the subspace model.

\section{Numerical results}
\label{section:Numerical results}
We examine noisy function fitting and flow matching. The function-fitting baseline uses Adam with finite-difference gradients, so both methods operate through function evaluations. The flow-matching experiment uses gradient-based Adam and compares it with standalone \texttt{2D-MoSub} and three hybrid configurations. We distinguish comparisons by function evaluations from comparisons by optimization steps: neither measure alone determines wall-clock training time.

\subsection{Noisy function fitting}
\label{section:Function Fitting}
The first task approximates the Ackley function,
\[
 f(\boldsymbol{x})=-20\exp\left(-0.2\sqrt{\frac{1}{d}\sum_{i=1}^d x_i^2}\right)
       -\exp\left(\frac{1}{d}\sum_{i=1}^d\cos(2\pi x_i)\right)+20+e,
\]
whose global minimum is at $\boldsymbol{x}=\boldsymbol{0}$. We use a four-layer multilayer perceptron (MLP) with 1,341 trainable parameters to fit samples corrupted by Gaussian noise with standard deviation $0.1$. The target function contains a sharp central minimum and oscillatory features. This is a regression problem: the optimization variables are the network parameters, and the loss landscape in parameter space should not be identified with the Ackley surface in input space.

\begin{figure}[htbp]
\centering
\includegraphics[width=\linewidth]{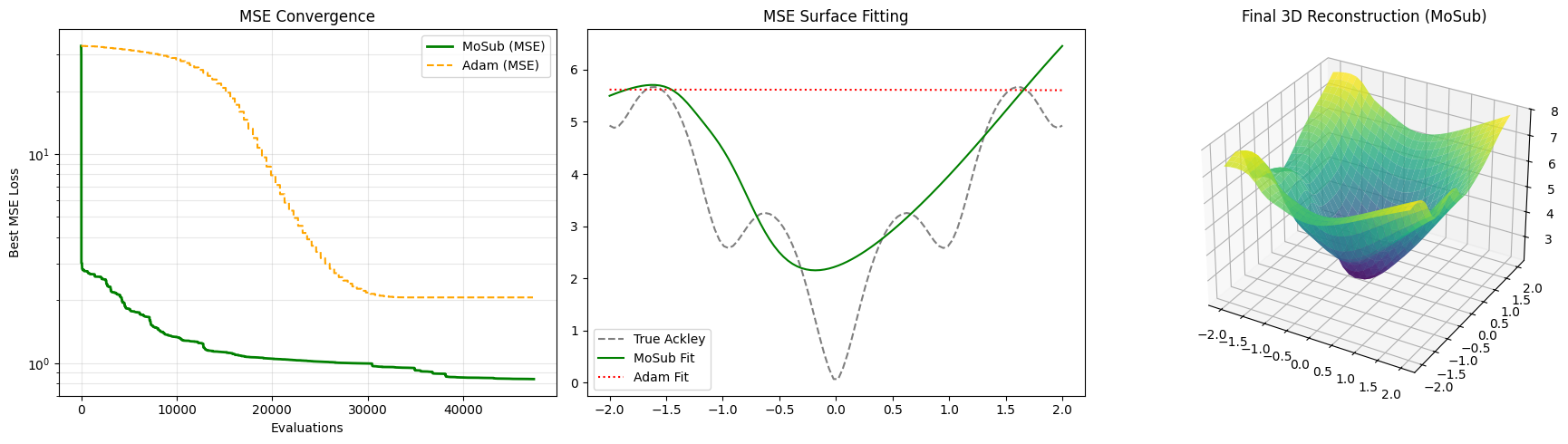}
\caption{Noisy Ackley function fitting with 1,341 trainable parameters. Left: best MSE versus function evaluations, with a logarithmic vertical axis. Middle: target and fitted functions along the slice $x_2=0$; the curve labeled Adam uses finite-difference gradients. Right: the surface reconstructed by \texttt{2D-MoSub}.}
\label{fig:3_Ackley}
\end{figure}

The training objective combines mean squared error (MSE) with an $L_2$ penalty,
\[
 \mathcal{L}(\boldsymbol{\theta})=\operatorname{MSE}(\boldsymbol{\theta})+\lambda\|\boldsymbol{\theta}\|_2^2,
 \qquad
 \operatorname{MSE}(\boldsymbol{\theta})=\frac1N\sum_{i=1}^N
 |\hat y_{\boldsymbol{\theta}}(\boldsymbol{x}_i)-y_i|^2,
 \qquad \lambda=10^{-4},
\]
where $N$ is the number of training samples. We compare \texttt{2D-MoSub} with numerical Adam, whose gradients are estimated by finite differences. Figure~\ref{fig:3_Ackley} reports MSE against function evaluations and shows the resulting fits.

We note that \texttt{2D-MoSub} reduces the MSE more rapidly and reaches a lower final value than numerical Adam. Its local quadratic models provide a way to use curvature information while keeping the interpolation problem small. The curves are consistent with an advantage in function-evaluation efficiency on this instance, although they do not isolate which algorithmic component accounts for that advantage.

The fitted slice provides a complementary view. \texttt{2D-MoSub} captures the broad central depression more clearly than the nearly flat Adam fit, but it still smooths the sharp minimum and misses several oscillatory features. The results therefore support an improvement in approximation quality relative to this baseline, rather than exact recovery of the Ackley function. They also do not establish escape from local minima in parameter space, which would require a separate analysis of the training objective.

\subsection{Learning a marginal vector field by flow matching}
\label{sec:flow}
The second task increases the parameter dimension and tests the hybrid strategies. We use a flow-matching model to learn the transport from a Gaussian initial distribution to a multimodal target distribution. The construction follows the conditional flow-matching framework of \cite{lipman2023flowmatching}; the required identities are stated below to define the training objective.

\subsubsection{Conditional and marginal training objectives}
A parameterized flow solves
\[
 \boldsymbol{X}_0\sim p_{\mathrm{init}},\qquad
 \frac{\mathrm{d}\boldsymbol{X}_t}{\mathrm{d}t}=\boldsymbol{u}_t^{\boldsymbol{\theta}}(\boldsymbol{X}_t),
\]
where the neural network $\boldsymbol{u}_t^{\boldsymbol{\theta}}$ is trained so that the distribution of $\boldsymbol{X}_t$ follows a prescribed probability path toward $p_{\mathrm{data}}$. Let $p_t(\boldsymbol{x}\mid \boldsymbol{z})$ and $\boldsymbol{u}_t^{\mathrm{target}}(\boldsymbol{x}\mid \boldsymbol{z})$ denote a conditional density and its generating vector field, given a target sample $\boldsymbol{z}$. Their marginal density is
\[
 p_t(\boldsymbol{x})=\int p_t(\boldsymbol{x}\mid \boldsymbol{z})p_{\mathrm{data}}(\boldsymbol{z})\,\mathrm{d}\boldsymbol{z}.
\]

\begin{theorem}[Marginalization; \cite{lipman2023flowmatching}]
\label{thm:marginalization}
Suppose the conditional density and vector field satisfy the continuity equation, with sufficient regularity to interchange integration and differentiation. Wherever $p_t(\boldsymbol{x})>0$, define
\[
 \boldsymbol{u}_t^{\mathrm{target}}(\boldsymbol{x})=\int \boldsymbol{u}_t^{\mathrm{target}}(\boldsymbol{x}\mid \boldsymbol{z})
 \frac{p_t(\boldsymbol{x}\mid \boldsymbol{z})p_{\mathrm{data}}(\boldsymbol{z})}{p_t(\boldsymbol{x})}\,\mathrm{d}\boldsymbol{z}.
\]
Then $p_t$ and $\boldsymbol{u}_t^{\mathrm{target}}$ satisfy the marginal continuity equation.
\end{theorem}

Direct regression onto the marginal field requires a potentially intractable integral. Conditional targets provide an alternative. With $t\sim\mathrm{U}(0,1)$, $\boldsymbol{z}\sim p_{\mathrm{data}}$, and $\boldsymbol{x}\sim p_t(\cdot\mid \boldsymbol{z})$, define
\begin{align}
 \mathcal{L}_{\mathrm{FM}}(\boldsymbol{\theta})
 &=\mathbb{E}\left[\|\boldsymbol{u}_t^{\boldsymbol{\theta}}(\boldsymbol{x})-\boldsymbol{u}_t^{\mathrm{target}}(\boldsymbol{x})\|_2^2\right],\\
 \mathcal{L}_{\mathrm{CFM}}(\boldsymbol{\theta})
 &=\mathbb{E}\left[\|\boldsymbol{u}_t^{\boldsymbol{\theta}}(\boldsymbol{x})-\boldsymbol{u}_t^{\mathrm{target}}(\boldsymbol{x}\mid \boldsymbol{z})\|_2^2\right].
\end{align}

\begin{theorem}[Equivalence of training objectives; \cite{lipman2023flowmatching}]
\label{thm:cfm}
If the relevant second moments are finite, then
\[
 \mathcal{L}_{\mathrm{CFM}}(\boldsymbol{\theta})=\mathcal{L}_{\mathrm{FM}}(\boldsymbol{\theta})+C,
 \qquad
 C=\mathbb{E}\left[\|\boldsymbol{u}_t^{\mathrm{target}}(\boldsymbol{x}\mid \boldsymbol{z})-\boldsymbol{u}_t^{\mathrm{target}}(\boldsymbol{x})\|_2^2\right],
\]
where $C\ge0$ is independent of $\boldsymbol{\theta}$. Their gradients coincide whenever differentiation under the expectation is justified.
\end{theorem}

To verify the identity, write the conditional target as its conditional expectation given $(t,\boldsymbol{x})$ plus a zero-mean residual. Expanding the squared error eliminates the cross term after conditioning on $(t,\boldsymbol{x})$, leaving the stated constant $C$. The same argument applies to a different fixed time-sampling distribution whenever its second moments are finite, with both losses and $C$ evaluated under that distribution. Conditional targets can therefore replace the marginalization integral for either gradient-based or derivative-free training under these assumptions.

\subsubsection{Gaussian paths and experimental configuration}
Take $p_{\mathrm{init}}=\mathcal{N}(\boldsymbol{0},\boldsymbol{I}_d)$ and define
\[
 \boldsymbol{x}_t=\alpha_t \boldsymbol{z}+\beta_t\boldsymbol{\epsilon},\qquad \boldsymbol{\epsilon}\sim\mathcal{N}(\boldsymbol{0},\boldsymbol{I}_d),
 \qquad p_t(\cdot\mid \boldsymbol{z})=\mathcal{N}(\alpha_t \boldsymbol{z},\beta_t^2 \boldsymbol{I}_d).
\]
The schedules satisfy $\alpha_0=0$, $\beta_0=1$, and approach $\alpha_1=1$, $\beta_1=0$ at the terminal time. For differentiable schedules with $\beta_t>0$, differentiating $\boldsymbol{x}_t$ gives
\begin{equation}
\label{eq:conditional-field}
 \boldsymbol{u}_t^{\mathrm{target}}(\boldsymbol{x}\mid \boldsymbol{z})
 =\left(\dot\alpha_t-\frac{\dot\beta_t}{\beta_t}\alpha_t\right)\boldsymbol{z}
 +\frac{\dot\beta_t}{\beta_t}\boldsymbol{x}.
\end{equation}
At a sampled point $\boldsymbol{x}_t$, the same target is $\boldsymbol{v}_t=\dot\alpha_t \boldsymbol{z}+\dot\beta_t\boldsymbol{\epsilon}$.

Here we use $\alpha_t=t$ and $\beta_t=\sqrt{1-t}$, rather than the linear schedule $\beta_t=1-t$. For independent $\boldsymbol{z}$ and $\boldsymbol{\epsilon}$, with $\mathbb{E}\|\boldsymbol{z}\|_2^2<\infty$, the sampled target and its second moment are
\begin{equation}
\label{eq:endpoint-moment}
 \boldsymbol{v}_t=\boldsymbol{z}-\frac{\boldsymbol{\epsilon}}{2\sqrt{1-t}},\qquad
 \mathbb{E}_{\boldsymbol{z},\boldsymbol{\epsilon}}\|\boldsymbol{v}_t\|_2^2
 =\mathbb{E}\|\boldsymbol{z}\|_2^2+\frac{d}{4(1-t)}.
\end{equation}
Integrating over $t\sim\mathrm{U}(0,1)$ gives an infinite second moment, even though any finite batch avoiding $t=1$ has finite targets almost surely. Thus, merely excluding the endpoint does not meet the finite-moment assumption of Theorem~\ref{thm:cfm}. One possible remedy is a specified truncation $t\sim\mathrm{U}(0,1-\delta)$ with $0<\delta<1$, for which
\begin{equation}
\label{eq:truncated-moment}
 \mathbb{E}\|\boldsymbol{v}_t\|_2^2
 =\mathbb{E}\|\boldsymbol{z}\|_2^2+
 \frac{d\log(1/\delta)}{4(1-\delta)}<\infty.
\end{equation}
This truncation ensures Theorem~\ref{thm:cfm} holds and the whole pipeline works. However, the linear schedule gives $\boldsymbol{v}_t=\boldsymbol{z}-\boldsymbol{\epsilon}$ and avoids this target singularity. We check that the linear schedule leads to a poor performance compared with $\beta_t=\sqrt{1-t}$.

\begin{algorithm}[htbp]
\caption{Conditional flow-matching loss for a Gaussian path}
\label{alg:flow-loss}
\begin{algorithmic}[1]
\Require Parameters $\boldsymbol{\theta}$, batch size $b$, schedules $\alpha_t,\beta_t$, and a specified time-sampling rule with valid targets.
\For{$i=1,\ldots,b$}
    \State Sample $\boldsymbol{z}_i\sim p_{\mathrm{data}}$, time $t_i$, and $\boldsymbol{\epsilon}_i\sim\mathcal{N}(\boldsymbol{0},\boldsymbol{I}_d)$.
    \State $\boldsymbol{x}_i\gets\alpha_{t_i}\boldsymbol{z}_i+\beta_{t_i}\boldsymbol{\epsilon}_i$.
    \State $\boldsymbol{v}_i\gets\dot\alpha_{t_i}\boldsymbol{z}_i+\dot\beta_{t_i}\boldsymbol{\epsilon}_i$.
\EndFor
\State \Return $b^{-1}\sum_{i=1}^b\|\boldsymbol{u}_{t_i}^{\boldsymbol{\theta}}(\boldsymbol{x}_i)-\boldsymbol{v}_i\|_2^2$.
\end{algorithmic}
\end{algorithm}

We use a four-layer MLP with 12,866 trainable parameters and compare five training configurations. The Adam learning rate is $10^{-3}$, and the reported horizon is 2,000 epochs/steps, following the plot's axis label. Table~\ref{tab:configurations} defines the hybrid periods using the figure legends. For subspace model construction, the sampled data, times, and Gaussian noise should remain fixed across the loss evaluations used to fit a single model, as specified in Algorithm~\ref{alg:interleaved}.

\begin{table}[htbp]
\centering
\caption{Flow-matching training configurations. The intervention periods follow the labels in Figure~\ref{fig:flow_comparison}.}
\label{tab:configurations}
\begin{tabular}{ll}
\toprule
Configuration & Update schedule\\
\midrule
Adam & Adam throughout\\
\texttt{2D-MoSub} & Derivative-free subspace search throughout\\
Hybrid 1 & One subspace update every $q=2$ steps\\
Hybrid 2 & One subspace update every $q=10$ steps\\
Hybrid 3 & Adam for the first half; \texttt{2D-MoSub} for the second half\\
\bottomrule
\end{tabular}
\end{table}

\subsubsection{Observed training behavior}
\begin{figure}[htbp]
\centering
\includegraphics[width=\textwidth]{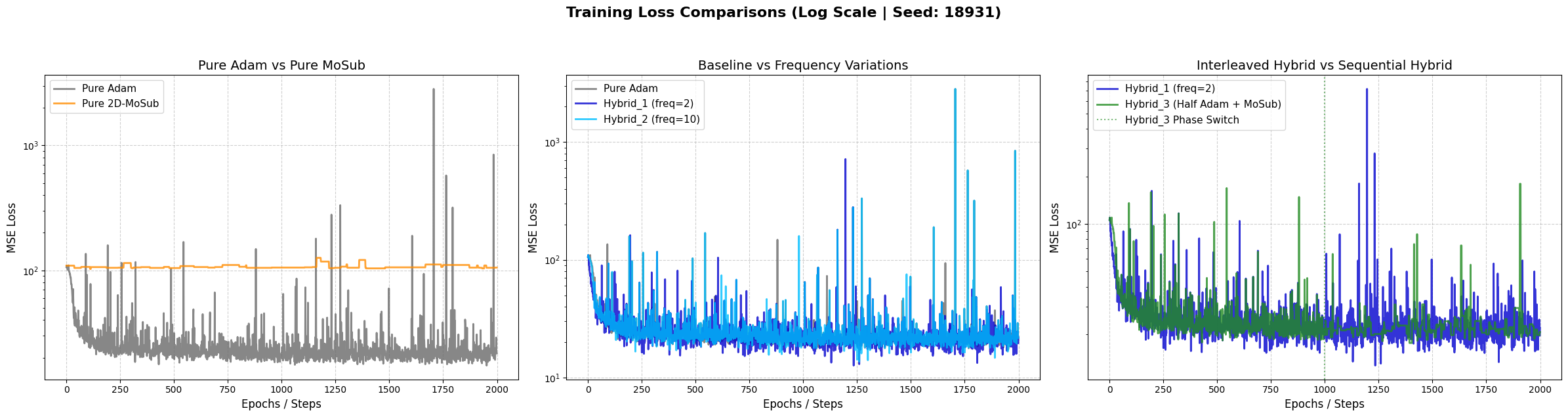}
\caption{Flow-matching training losses over 2,000 epochs/steps, with Adam learning rate $10^{-3}$ and seed 18931 as indicated in the plot. Left: Adam and standalone \texttt{2D-MoSub}. Middle: Adam and the two interleaved hybrids. Right: Hybrid 1 and the sequential Hybrid 3, whose phase switch occurs halfway through training. The vertical axes are logarithmic; the curves display substantial fluctuations.}
\label{fig:flow_comparison}
\end{figure}

Figure~\ref{fig:flow_comparison} shows that standalone \texttt{2D-MoSub} makes little progress in this example, whereas Adam and the hybrid configurations reduce the loss. We check that the final trust-region radius is $5\times10^{-7}$, which is consistent with an ineffective local search at initialization. A small radius alone, however, does not establish that the iterate is a local minimum, and the loss curves do not identify the precise cause of stagnation.

The interleaved methods reach loss levels similar to Adam, but all retain noticeable fluctuations and occasional spikes. The plotted run therefore supports the feasibility of combining Adam with subspace updates, without establishing a consistent reduction in oscillations. Hybrid 3 maintains a similar loss scale after switching to \texttt{2D-MoSub}. This suggests that derivative-free refinement can remain useful after a gradient-based initialization, although the figure does not establish superiority in final accuracy or training time.

\begin{figure}[p]
\centering
\includegraphics[width=0.95\textwidth]{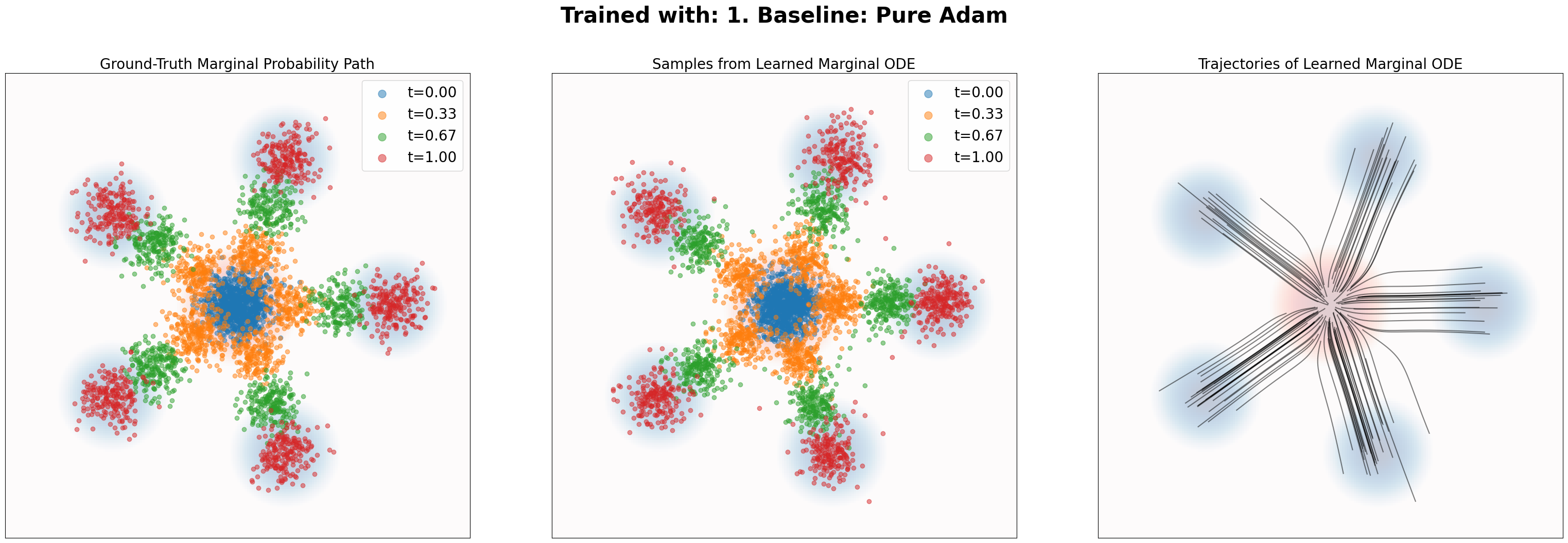}
\par\vspace{0.6em}
\includegraphics[width=0.95\textwidth]{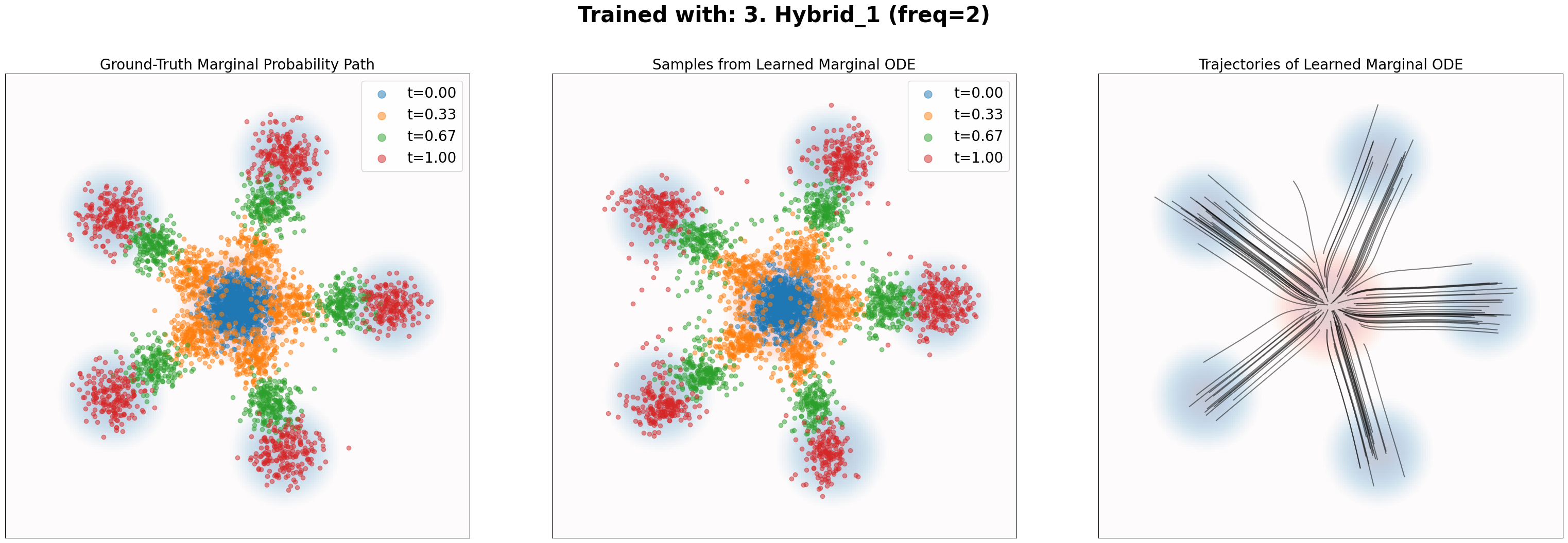}
\par\vspace{0.6em}
\includegraphics[width=0.95\textwidth]{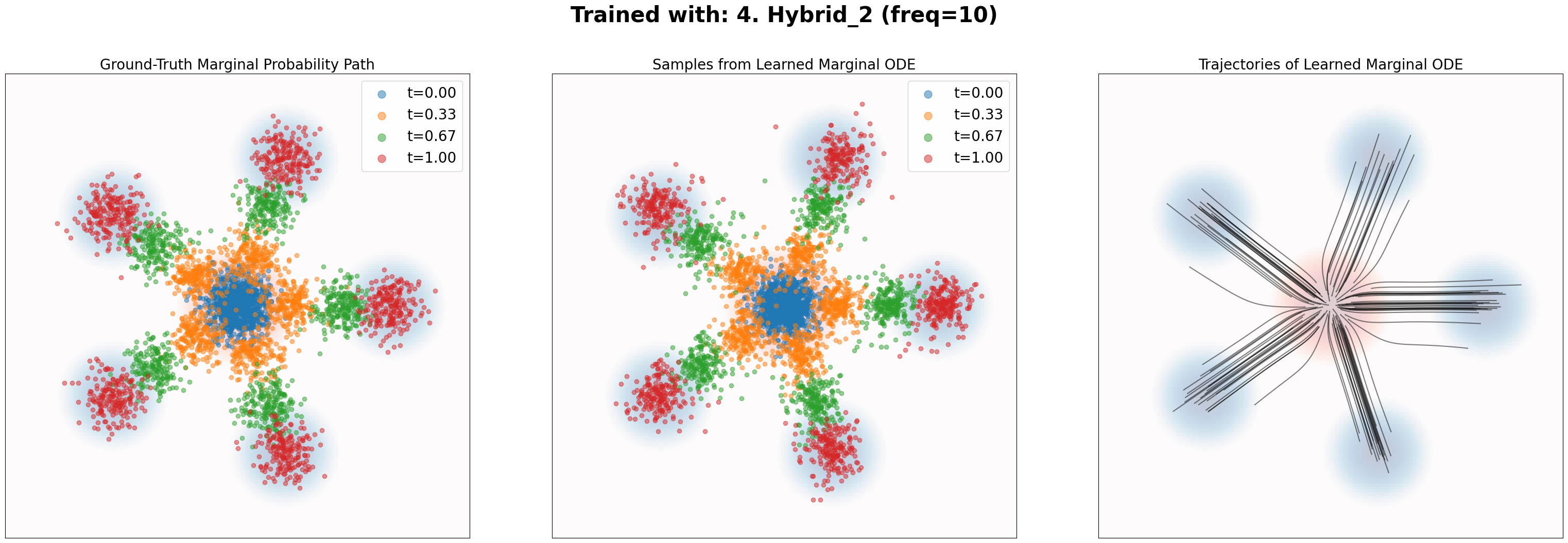}
\par\vspace{0.6em}
\includegraphics[width=0.95\textwidth]{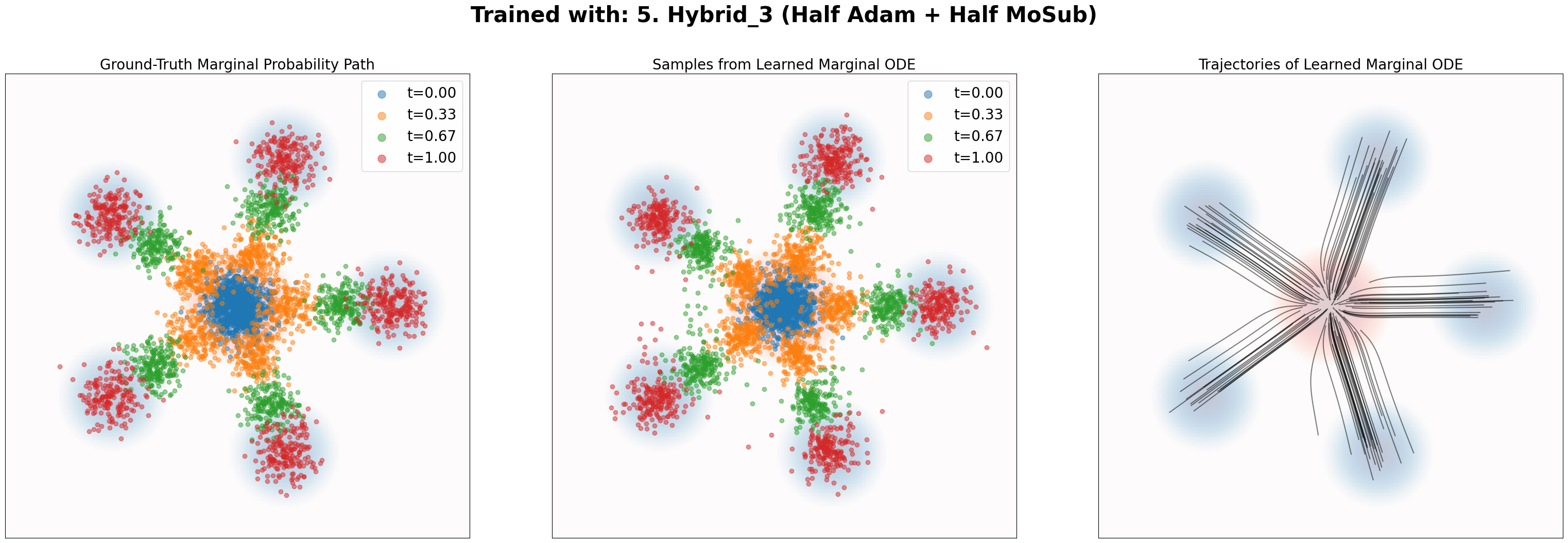}
\caption{Flow-matching visualizations for Adam, Hybrid 1, Hybrid 2, and Hybrid 3, from top to bottom. Each row shows reference marginal samples (left), samples generated by the learned ordinary differential equation (middle), and learned trajectories (right). Point colors indicate time, from blue at $t=0$ to red at $t=1$. In the trajectory panels, the central red shading marks the initial distribution and the blue shading marks the target modes.}
\label{fig:flow_matching_output}
\end{figure}

Figure~\ref{fig:flow_matching_output} shows that the trained models recover the broad multimodal structure and transport samples from the central initial distribution toward the target modes. These visual comparisons are qualitative. Quantitative distributional metrics and repeated runs would be needed to determine whether one optimizer yields consistently more accurate probability paths or better sample quality.

\subsection{Interpretation and experimental limitations}
The experiments address different comparisons. Ackley fitting tests two function-value-based optimizers on one regression instance; flow matching tests whether gradient assistance improves on standalone subspace search in a larger model. The implementation details of the experimental studies are available
in \url{https://github.com/DMwang5/2D-Mosub-neural-network-numerical-result-}.

The main limitation of our work is the lack of GPU, which means we could only work on a limited scale of network to solve the instances in a reasonable time. However, we believe that our algorithm could fit the large-scale network optimization. The potential risk lies in the trial step of \texttt{2D-MoSub}. Specifically, since the initial direction of \texttt{2D-MoSub} is picked randomly, we are more likely to get trapped with increasing trainable parameters. In order to tackle this problem, the hybrid methods introduce Adam to provide a valuable guidance.

\section{Theoretical analysis}
\label{sec:theory}
This section separates model estimation from optimization guarantees. We first analyze the mean and covariance of a quadratic model fitted on geometry chosen independently of its batch. We then characterize exact-loss acceptance and distinguish scalar loss error from full-gradient error. Finally, we analyze hybrids with stochastic gradient steps under randomized and deterministic schedules. Throughout, $n$ denotes the number of trainable parameters, $d$ the data-space dimension, $s$ the fixed dataset size, and $b$ and $B$ the loss and gradient batch sizes. Smoothness and variance bounds are assumptions on the stated objective; they are not inferred from the network parameter count or the plotted curves.

\subsection{Unbiasedness with fixed interpolation geometry}
Let $\mathcal{S}=\{\boldsymbol{z}_1,\ldots,\boldsymbol{z}_s\}$ be a fixed dataset and define
\[
 \mathcal{L}(\boldsymbol{\theta})=\frac1s\sum_{i=1}^s\ell(\boldsymbol{\theta},\boldsymbol{z}_i).
\]
Draw indices $I_1,\ldots,I_b$ independently and uniformly from $\{1,\ldots,s\}$, and set
\begin{equation}
\label{eq:batch-loss}
 \widehat{\mathcal{L}}_b(\boldsymbol{\theta})=\frac1b\sum_{r=1}^b\ell(\boldsymbol{\theta},\boldsymbol{z}_{I_r}).
\end{equation}
Then $\mathbb{E}[\widehat{\mathcal{L}}_b(\boldsymbol{\theta})]=\mathcal{L}(\boldsymbol{\theta})$ for fixed $\boldsymbol{\theta}$. Here $b$ is the batch size, rather than a ratio between dataset and batch sizes.

Fix a base point $\bar{\boldsymbol{\theta}}$, an orthonormal matrix $\boldsymbol{P}\in\mathbb{R}^{n\times2}$, and coordinates $Y=\{\boldsymbol{y}_1,\ldots,\boldsymbol{y}_m\}\subset\mathbb{R}^2$. The corresponding parameter-space points are $\bar{\boldsymbol{\theta}}+\boldsymbol{P}\boldsymbol{y}_j$. Define
\[
 \boldsymbol{\phi}(\boldsymbol{y})=(1,y_1,y_2,\tfrac12 y_1^2,y_1y_2,\tfrac12 y_2^2)^T,
 \qquad \boldsymbol{\Phi}_{j,:}=\boldsymbol{\phi}(\boldsymbol{y}_j)^T.
\]
Assume $m\ge6$ and $\boldsymbol{\Phi}\in\mathbb{R}^{m\times6}$ has full column rank. Write
\[
 F_j=\mathcal{L}(\bar{\boldsymbol{\theta}}+\boldsymbol{P}\boldsymbol{y}_j),\qquad
 \widehat F_j=\widehat{\mathcal{L}}_b(\bar{\boldsymbol{\theta}}+\boldsymbol{P}\boldsymbol{y}_j),\qquad
 \boldsymbol{w}^*=\boldsymbol{\Phi}^\dagger \boldsymbol{F},\qquad \widehat{\boldsymbol{w}}=\boldsymbol{\Phi}^\dagger\widehat{\boldsymbol{F}}.
\]
The models are $m^*(\boldsymbol{y})=\boldsymbol{\phi}(\boldsymbol{y})^T\boldsymbol{w}^*$ and $\widehat m(\boldsymbol{y})=\boldsymbol{\phi}(\boldsymbol{y})^T\widehat{\boldsymbol{w}}$. The reference model $m^*$ is fitted to full-data losses; it need not equal the objective or its Taylor polynomial.

\begin{proposition}
\label{prop:unbiased}
Suppose the sampled losses are integrable and the geometry $(\bar{\boldsymbol{\theta}},\boldsymbol{P},Y)$ is fixed independently of the batch used to evaluate $\widehat{\boldsymbol{F}}$. Then
\[
 \mathbb{E}[\widehat{\boldsymbol{w}}\mid\bar{\boldsymbol{\theta}},\boldsymbol{P},Y]=\boldsymbol{w}^*,\qquad
 \mathbb{E}[\widehat m(\boldsymbol{y})\mid\bar{\boldsymbol{\theta}},\boldsymbol{P},Y]=m^*(\boldsymbol{y})
\]
for every fixed $\boldsymbol{y}\in\mathbb{R}^2$. This covers exact interpolation when $m=6$ and ordinary least squares when $m>6$.
\end{proposition}
\begin{proof}
Conditioning on the geometry leaves $\boldsymbol{\Phi}^\dagger$ deterministic. Equation~\eqref{eq:batch-loss} gives $\mathbb{E}[\widehat{\boldsymbol{F}}\mid\bar{\boldsymbol{\theta}},\boldsymbol{P},Y]=\boldsymbol{F}$, so
\[
 \mathbb{E}[\widehat{\boldsymbol{w}}\mid\bar{\boldsymbol{\theta}},\boldsymbol{P},Y]
 =\boldsymbol{\Phi}^\dagger\mathbb{E}[\widehat{\boldsymbol{F}}\mid\bar{\boldsymbol{\theta}},\boldsymbol{P},Y]
 =\boldsymbol{\Phi}^\dagger \boldsymbol{F}=\boldsymbol{w}^*.
\]
Multiplication by $\boldsymbol{\phi}(\boldsymbol{y})^T$ proves the model identity.
\end{proof}

Proposition~\ref{prop:unbiased} also applies conditionally on the optimization history if the geometry is chosen from that history and a fresh batch is then sampled. It should be noticed that unbiasedness alone does not control the model error: poor conditioning of $\boldsymbol{\Phi}$ can amplify noise, and neither the minimizer of the stochastic model nor its predicted reduction is generally unbiased. We shall discuss this amplification in the following.

\begin{proposition}
\label{prop:model-covariance}
Under the assumptions of Proposition~\ref{prop:unbiased}, suppose each single-sample evaluation vector
\[
 \boldsymbol{a}(\boldsymbol{z})=\big(\ell(\bar{\boldsymbol{\theta}}+\boldsymbol{P}\boldsymbol{y}_1,\boldsymbol{z}),\ldots,
                 \ell(\bar{\boldsymbol{\theta}}+\boldsymbol{P}\boldsymbol{y}_m,\boldsymbol{z})\big)^T
\]
has finite second moments. Conditional on the geometry, let $\boldsymbol{\Sigma}=\operatorname{Cov}_{\boldsymbol{z}}(\boldsymbol{a}(\boldsymbol{z}))$ and $\boldsymbol{A}=\boldsymbol{\Phi}^\dagger$. If the same batch of $b$ independent samples is used at all $m$ points, then
\begin{align}
 \operatorname{Cov}(\widehat{\boldsymbol{w}}\mid\bar{\boldsymbol{\theta}},\boldsymbol{P},Y)
 &=\frac1b \boldsymbol{A}\boldsymbol{\Sigma} \boldsymbol{A}^T,\label{eq:model-covariance}\\
 \mathbb{E}[\|\widehat{\boldsymbol{w}}-\boldsymbol{w}^*\|_2^2\mid\bar{\boldsymbol{\theta}},\boldsymbol{P},Y]
 &=\frac1b\operatorname{tr}(\boldsymbol{A}\boldsymbol{\Sigma} \boldsymbol{A}^T)
 \le\frac{\operatorname{tr}(\boldsymbol{\Sigma})}{b\,\sigma_{\min}(\boldsymbol{\Phi})^2}.
 \label{eq:model-noise-bound}
\end{align}
For fixed $\boldsymbol{y}$, the model-value MSE is
\begin{equation}
\label{eq:model-value-variance}
 \mathbb{E}[|\widehat m(\boldsymbol{y})-m^*(\boldsymbol{y})|^2\mid\bar{\boldsymbol{\theta}},\boldsymbol{P},Y]
 =\frac1b\boldsymbol{\phi}(\boldsymbol{y})^T\boldsymbol{A}\boldsymbol{\Sigma} \boldsymbol{A}^T\boldsymbol{\phi}(\boldsymbol{y}).
\end{equation}
\end{proposition}
\begin{proof}
The evaluation vector satisfies $\widehat{\boldsymbol{F}}=b^{-1}\sum_{r=1}^b \boldsymbol{a}(\boldsymbol{z}_{I_r})$, so its conditional covariance is $\boldsymbol{\Sigma}/b$. Applying the linear map $\boldsymbol{A}$ gives \eqref{eq:model-covariance}. Unbiasedness converts covariance traces into mean squared errors. Finally,
$\operatorname{tr}(\boldsymbol{A}\boldsymbol{\Sigma} \boldsymbol{A}^T)\le\|\boldsymbol{A}\|_2^2\operatorname{tr}(\boldsymbol{\Sigma})$ and $\|\boldsymbol{A}\|_2=1/\sigma_{\min}(\boldsymbol{\Phi})$ give the bound; the scalar identity follows by applying $\boldsymbol{\phi}(\boldsymbol{y})^T$.
\end{proof}

The shared batch generally correlates errors at different interpolation points; these correlations are retained in $\boldsymbol{\Sigma}$. For two points, for example, the noise variance in their loss difference equals
\[
 \frac1b\big(\Sigma_{ii}+\Sigma_{jj}-2\Sigma_{ij}\big).
\]
Positive covariance can therefore improve the accuracy of a loss difference, although batch reuse alone does not ensure positive covariance or an accurate model. Moreover, the bounds concern sampling error relative to $m^*$, not the approximation error $m^*(\boldsymbol{y})-\mathcal{L}(\bar{\boldsymbol{\theta}}+\boldsymbol{P}\boldsymbol{y})$.

\subsection{Convergence of safeguarded loss values}
\begin{proposition}
\label{prop:rollback}
Suppose $\mathcal{L}$ is bounded below and finite at $\boldsymbol{\theta}_0$. Given any possibly random proposal $\boldsymbol{\theta}_{\mathrm{trial}}$, update by
\[
 \boldsymbol{\theta}_{k+1}=\begin{cases}
 \boldsymbol{\theta}_{\mathrm{trial}},&\mathcal{L}(\boldsymbol{\theta}_{\mathrm{trial}})<\mathcal{L}(\boldsymbol{\theta}_k),\\
 \boldsymbol{\theta}_k,&\text{otherwise}.
 \end{cases}
\]
If this exact-loss rule is applied at every iteration, then $\{\mathcal{L}(\boldsymbol{\theta}_k)\}$ is non-increasing and converges to a finite limit for every realization of the proposals.
\end{proposition}
\begin{proof}
The acceptance rule gives $\mathcal{L}(\boldsymbol{\theta}_{k+1})\le\mathcal{L}(\boldsymbol{\theta}_k)$ at each iteration. Every non-increasing real sequence bounded below has a finite limit.
\end{proof}

This is a statement about objective values. It does not imply convergence of the iterates, stationarity, or convergence to a global minimum. In Algorithm~\ref{alg:interleaved}, the rule is applied only to subspace proposals, so it guarantees non-increase only across those updates. If acceptance uses a sampled loss, the same conclusion holds for a fixed acceptance objective, but need not hold for the full loss or for independently resampled objectives.

\subsection{Sampling errors in loss and gradient estimates}
A scalar loss estimate and a full gradient estimate have different mean squared errors. The following comparison makes the norms and dimensional assumptions explicit.

\begin{proposition}
\label{prop:variance}
At a fixed $\boldsymbol{\theta}\in\mathbb{R}^n$, assume the sampled losses are differentiable and
\[
 \operatorname{Var}_{\boldsymbol{z}}[\ell(\boldsymbol{\theta},\boldsymbol{z})]=\sigma_\ell^2<\infty,
 \qquad
 \frac1n\sum_{j=1}^n\operatorname{Var}_{\boldsymbol{z}}[\nabla_j\ell(\boldsymbol{\theta},\boldsymbol{z})]
 =\sigma_g^2<\infty.
\]
For independent uniform samples, let $\widehat{\mathcal{L}}_b$ be the batch loss and let $\widehat{\boldsymbol{g}}_B$ be the average of $B$ sample gradients. Then
\begin{align}
 \mathbb{E}\left[|\widehat{\mathcal{L}}_b(\boldsymbol{\theta})-\mathcal{L}(\boldsymbol{\theta})|^2\right]
 &=\frac{\sigma_\ell^2}{b},\label{eq:loss-variance}\\
 \mathbb{E}\left[\|\widehat{\boldsymbol{g}}_B(\boldsymbol{\theta})-\boldsymbol{\nabla}\mathcal{L}(\boldsymbol{\theta})\|_2^2\right]
 &=\frac{n\sigma_g^2}{B}.\label{eq:gradient-variance}
\end{align}
Thus, absolute MSE at most $\varepsilon^2$ is ensured by
\[
 b\ge\max\{1,\lceil\sigma_\ell^2/\varepsilon^2\rceil\},\qquad
 B\ge\max\{1,\lceil n\sigma_g^2/\varepsilon^2\rceil\}.
\]
\end{proposition}
\begin{proof}
Independence makes the variance of a sample average equal to the single-sample variance divided by the batch size. For the vector estimate, the expected squared Euclidean error is the sum of the coordinate variances, which gives \eqref{eq:gradient-variance}.
\end{proof}

For fixed accuracy and dimension-independent variance bounds, these sufficient batch sizes scale as $O(1)$ for the scalar loss and $O(n)$ for the full gradient. The comparison uses absolute scalar error and total vector error; it is not an equal signal-to-noise or equal optimization-progress comparison. In particular, Proposition~\ref{prop:variance} establishes an $O(1/n)$ ratio of training samples under the same variance bounds. 

\subsection{A stationarity bound for an idealized hybrid}
We next analyze a randomized scheme that takes an unbiased stochastic gradient step with probability $\alpha\in(0,1]$ and a safeguarded subspace step otherwise. The subspace step may use any proposal mechanism, provided its acceptance test never increases the exact objective.

\begin{proposition}
\label{prop:complexity}
Suppose $\mathcal{L}$ is $L$-smooth and bounded below by $\mathcal{L}_{\min}$. At each iteration, conditionally on the history $\mathcal{H}_k$, choose a first-order step with probability $\alpha$, and use
\[
 \boldsymbol{\theta}_{k+1}=\boldsymbol{\theta}_k-h\widehat{\boldsymbol{g}}_B(\boldsymbol{\theta}_k),\qquad 0<h\le1/L,
\]
where the gradient batch is sampled independently of the update-type selection, conditional on $\mathcal{H}_k$, and
\[
 \mathbb{E}[\widehat{\boldsymbol{g}}_B(\boldsymbol{\theta}_k)\mid\mathcal{H}_k]
 =\boldsymbol{\nabla}\mathcal{L}(\boldsymbol{\theta}_k),\qquad
 \mathbb{E}[\|\widehat{\boldsymbol{g}}_B(\boldsymbol{\theta}_k)-\boldsymbol{\nabla}\mathcal{L}(\boldsymbol{\theta}_k)\|_2^2
 \mid\mathcal{H}_k]\le\sigma^2/B.
\]
Otherwise, take a subspace step satisfying $\mathcal{L}(\boldsymbol{\theta}_{k+1})\le\mathcal{L}(\boldsymbol{\theta}_k)$ almost surely. Choose $J$ uniformly from $\{0,\ldots,T-1\}$, independently of the trajectory, and set $\boldsymbol{\theta}_{\mathrm{out}}=\boldsymbol{\theta}_J$. Then
\begin{equation}
\label{eq:stationarity}
 \mathbb{E}[\|\boldsymbol{\nabla}\mathcal{L}(\boldsymbol{\theta}_{\mathrm{out}})\|_2^2]
 \le\frac{2D}{\alpha hT}+\frac{Lh\sigma^2}{B},
 \qquad D=\mathcal{L}(\boldsymbol{\theta}_0)-\mathcal{L}_{\min}.
\end{equation}
In particular, choosing
\[
 B\ge\max\{1,\lceil 2Lh\sigma^2/\varepsilon^2\rceil\},\qquad
 T\ge\max\{1,\lceil4D/(\alpha h\varepsilon^2)\rceil\}
\]
ensures $\mathbb{E}[\|\boldsymbol{\nabla}\mathcal{L}(\boldsymbol{\theta}_{\mathrm{out}})\|_2^2]\le\varepsilon^2$.
\end{proposition}
\begin{proof}
By the $L$-smoothness, the parameter update $\boldsymbol{\theta}_{k+1}=\boldsymbol{\theta}_k-h\widehat{\boldsymbol{g}}_B(\boldsymbol{\theta}_k)$ yields an upper bound:
    \begin{align*}
        \mathcal{L}(\boldsymbol{\theta}_{k+1})\le \mathcal{L}(\boldsymbol{\theta}_k)+\nabla\mathcal{L}(\boldsymbol{\theta}_k)^T(\boldsymbol{\theta}_{k+1}-\boldsymbol{\theta}_k)+\frac{L}{2}||\boldsymbol{\theta}_{k+1}-\boldsymbol{\theta}_k||^2.
    \end{align*}
By smoothness and conditional unbiasedness, a first-order update satisfies
\begin{align*}
 \mathbb{E}[\mathcal{L}(\boldsymbol{\theta}_{k+1})\mid\mathcal{H}_k,\mathrm{FO}]
 &\le\mathcal{L}(\boldsymbol{\theta}_k)
 -\left(h-\frac{Lh^2}{2}\right)\|\boldsymbol{\nabla}\mathcal{L}(\boldsymbol{\theta}_k)\|_2^2
 +\frac{Lh^2\sigma^2}{2B}\\
 &\le\mathcal{L}(\boldsymbol{\theta}_k)-\frac h2\|\boldsymbol{\nabla}\mathcal{L}(\boldsymbol{\theta}_k)\|_2^2
 +\frac{Lh^2\sigma^2}{2B}.
\end{align*}
where the last inequality comes from the learning rate condition. Define the indicator variable $\mathcal{I}_k=\{0,1\}$, where $\mathcal{I}_k=1$ if the subspace search yields a strictly lower loss. Then the subspace update satisfies 
    \begin{align*}
        \mathbb{E}[\mathcal{L}(\boldsymbol{\theta}_{k+1})\mid\mathcal{H}_k,\mathrm{ZO}]=\mathcal{L}(\boldsymbol{\theta}_k)-\mathbb{E}[\mathcal{I}_k\cdot\mathcal{F}_{\mathrm{sub}}(\boldsymbol{\theta}_k)]
    \end{align*}
    Since $\mathbb{E}[\mathcal{I}_k\cdot\mathcal{F}_{\mathrm{sub}}(\theta_k)]$ is non-negative, the ZO step is strictly non-increasing and potentially advantageous. Averaging over the update type therefore gives
\[
 \mathbb{E}[\mathcal{L}(\boldsymbol{\theta}_{k+1})\mid\mathcal{H}_k]
 \le\mathcal{L}(\boldsymbol{\theta}_k)-\frac{\alpha h}{2}\|\boldsymbol{\nabla}\mathcal{L}(\boldsymbol{\theta}_k)\|_2^2
 +\frac{\alpha Lh^2\sigma^2}{2B}.
\]
Taking total expectations, summing over $k$, and using the lower bound yields
\[
 \frac{\alpha h}{2}\sum_{k=0}^{T-1}
 \mathbb{E}[\|\boldsymbol{\nabla}\mathcal{L}(\boldsymbol{\theta}_k)\|_2^2]
 \le D+\frac{T\alpha Lh^2\sigma^2}{2B}.
\]
Division by $\alpha hT/2$ proves \eqref{eq:stationarity}. The stated choices of $B$ and $T$ bound each term on its right-hand side by $\varepsilon^2/2$.
\end{proof}

The noise term $Lh\sigma^2/B$ is essential: increasing the number of iterations alone does not make this bound arbitrarily small when the step size and batch size are fixed. The case $\alpha=1$ is stochastic gradient descent, while $\alpha=0$ is outside the result. Proving stationarity for pure subspace search would require additional assumptions. Next we shall discuss the expected stationarity of deterministic schedules.

\begin{proposition}
\label{prop:deterministic}
Let $\mathcal{I}\subseteq\{0,\ldots,T-1\}$ be a deterministic set of stochastic-gradient update indices, with $N_{\mathrm{FO}}=|\mathcal{I}|\ge1$. Assume the smoothness, lower bound, step size, and conditional gradient moment bounds of Proposition~\ref{prop:complexity}. At all remaining indices, use exact-loss safeguarded subspace updates. If $J$ is uniform on $\mathcal{I}$ and independent of the trajectory, then
\begin{equation}
\label{eq:deterministic-stationarity}
 \mathbb{E}\|\boldsymbol{\nabla}\mathcal{L}(\boldsymbol{\theta}_J)\|_2^2
 \le \frac{2D}{hN_{\mathrm{FO}}}+\frac{Lh\sigma^2}{B}.
\end{equation}
\end{proposition}
\begin{proof}
Use the first-order descent inequality from Proposition~\ref{prop:complexity} at each $k\in\mathcal{I}$ and non-increase at all other indices. Taking expectations and summing gives
\[
 \frac h2\sum_{k\in\mathcal{I}}\mathbb{E}\|\boldsymbol{\nabla}\mathcal{L}(\boldsymbol{\theta}_k)\|_2^2
 \le D+\frac{N_{\mathrm{FO}}Lh^2\sigma^2}{2B}.
\]
Divide by $hN_{\mathrm{FO}}/2$ and use the distribution of $J$.
\end{proof}

For the periodic schedule, $N_{\mathrm{FO}}=T-\lfloor T/q\rfloor$; for a sequential schedule with $K_{\mathrm{A}}\ge1$ gradient updates, $N_{\mathrm{FO}}=K_{\mathrm{A}}$. This connects the analysis to the two scheduling patterns. The output is selected from iterates immediately before gradient updates. In the sequential case, the guarantee therefore concerns the first phase; monotone subspace refinement does not by itself preserve a small gradient norm or give a stationarity guarantee for the final iterate.

To describe computational cost, suppose each subspace update uses at most $r$ batch-loss evaluations with batch size $b$ and at most $c$ full-data loss evaluations on a dataset of size $s$. Counting a sample used in a loss or gradient evaluation as one sample evaluation, the expected cost over $T$ iterations is bounded by
\begin{equation}
\label{eq:sample-cost}
 S_{\mathrm{total}}\le
 T\big[\alpha B+(1-\alpha)(rb+cs)\big].
\end{equation}
With the choices in Proposition~\ref{prop:complexity}, this gives, up to integer rounding,
\[
 S_{\mathrm{total}}=
 O\!\left(\frac{D}{h\varepsilon^2}
 \left[B+\frac{1-\alpha}{\alpha}(rb+cs)\right]\right).
\]

\section{Conclusion and future work}
\label{sec:conclusion}
We investigated \texttt{2D-MoSub} for neural network optimization, both as a standalone method and in combination with Adam. In the noisy Ackley fitting experiment, the subspace method achieved a lower MSE than the finite-difference Adam baseline. In the flow-matching experiment, standalone subspace search stagnated early, whereas the gradient-assisted configurations reduced the loss. These observations motivate using subspace searches selectively and assessing their contribution under explicit evaluation budgets.

The analysis establishes conditional unbiasedness and quantifies sampling-noise amplification in fixed-geometry quadratic models. It also proves convergence of safeguarded loss values and stationarity bounds for randomized and deterministic hybrids with stochastic gradient updates. These results identify requirements on geometry, sampling, and acceptance; they do not prove convergence of the hybrid methods. In particular, monotone refinement of the loss is insufficient to guarantee stationarity of the final network.

Further work should examine adaptive intervention schedules, interpolation geometry under stochastic noise, and acceptance tests that reduce the cost of full-data evaluations. For flow matching, documenting and controlling the terminal-time behavior is also necessary. Broader experiments with repeated runs, distributional error metrics, and measured training time are needed to assess when hybrid subspace training offers a practical advantage.

\section*{Funding}
This work was partially supported by Laboratory Directed Research and Development (LDRD) funding from Lawrence Berkeley National Laboratory and by the U.S. Department of Energy, Office of Science, Office of Advanced Scientific Computing Research applied mathematics program (SEAZOTIE) under Contract Number DE-AC02-05CH11231.

\bibliographystyle{unsrtnat}
\bibliography{churui_refs}

\appendix
\section{Overview of \texttt{2D-MoSub}}
\label{section:2D-MoSub}
The two-dimensional model-based subspace method of \cite{xie2023twodimensional} solves unconstrained problems $\min_{\boldsymbol{x}\in\mathbb{R}^n}f(\boldsymbol{x})$ using quadratic models in affine subspaces. The outline below summarizes the features relevant to this paper. It is not a replacement for the interpolation geometry, modified-model construction, and trust-region update rules in the original method.

At iteration $k$, let $\boldsymbol{d}_1^{(k)}$ and $\boldsymbol{d}_2^{(k)}$ be orthonormal directions, and write $\boldsymbol{P}_k=[\boldsymbol{d}_1^{(k)},\boldsymbol{d}_2^{(k)}]$. A quadratic $q_k(\boldsymbol{y})$ approximates $f(\boldsymbol{x}_k+\boldsymbol{P}_k\boldsymbol{y})$ for $\boldsymbol{y}\in\mathbb{R}^2$. The trust-region search seeks a trial coordinate $\boldsymbol{y}_k^+$ within $\|\boldsymbol{y}\|_2\le\Delta_k$, giving $\boldsymbol{x}_k^+=\boldsymbol{x}_k+\boldsymbol{P}_k\boldsymbol{y}_k^+$. For positive predicted reduction, the agreement ratio is
\begin{equation}
\label{eq:trust-ratio}
 \rho_k=\frac{f(\boldsymbol{x}_k)-f(\boldsymbol{x}_k^+)}{q_k(\boldsymbol{0})-q_k(\boldsymbol{y}_k^+)}.
\end{equation}
This ratio guides the acceptance of the trial point and the update of the trust-region radius. The original method also allows a modified quadratic model in the trial-step construction.

\begin{algorithm}[htbp]
\caption{High-level outline of \texttt{2D-MoSub} \cite{xie2023twodimensional}}
\label{alg:mosub}
\begin{algorithmic}[1]
\Require Initial point $\boldsymbol{x}_0$, initial unit direction $\boldsymbol{d}_1^{(0)}$, radius $\Delta_0>0$, termination tolerance, and trust-region parameters.
\State Initialize the interpolation data and a one-dimensional quadratic model along $\boldsymbol{d}_1^{(0)}$.
\For{$k=0,1,\ldots$ until termination}
    \State Choose a unit direction $\boldsymbol{d}_2^{(k)}$ orthogonal to $\boldsymbol{d}_1^{(k)}$.
    \State Construct or update interpolation points in $\boldsymbol{x}_k+\operatorname{span}\{\boldsymbol{d}_1^{(k)},\boldsymbol{d}_2^{(k)}\}$.
    \State Construct the two-dimensional quadratic model $q_k$.
    \State Compute a trust-region trial point, using the modified-model procedure when required.
    \State Evaluate the objective and use model agreement to accept or reject the trial and update the radius.
    \If{$\boldsymbol{x}_{k+1}\ne \boldsymbol{x}_k$}
        \State $\boldsymbol{d}_1^{(k+1)}\gets(\boldsymbol{x}_{k+1}-\boldsymbol{x}_k)/\|\boldsymbol{x}_{k+1}-\boldsymbol{x}_k\|_2$.
    \Else
        \State Retain a valid unit direction for the next subspace.
    \EndIf
    \State Retain the one-dimensional model information needed for the next iteration.
\EndFor
\end{algorithmic}
\end{algorithm}

The direction update must handle a rejected or zero-length step explicitly; normalizing $\boldsymbol{x}_{k+1}-\boldsymbol{x}_k$ is undefined in that case. Likewise, the ratio in \eqref{eq:trust-ratio} is used only when the predicted reduction is positive. In the hybrid setting, the full-loss acceptance test in Algorithm~\ref{alg:interleaved} provides an additional check on a proposal obtained from a stochastic subspace model.

\end{document}